# Mechanistic Interpretability-Guided Selective Fine-Tuning of Vision–Language Models for Centimeter-Level Flood Depth Estimation

Nafis Fuad
*Department. of Civil & Environmental Engineering*
*Wayne State University*
Detroit, MI, USA
nafisfuad@wayne.edu

Xiaodong Qian
*Department of Civil & Environmental Engineering*
*Institute for AI and Data Science (AIDaS)*
*Wayne State University*
Detroit, MI, USA
xdqian@wayne.edu

Dongxiao Zhu
*Department of Computer Science,*
*Institute for AI and Data Science (AIDaS)*
*Wayne State University*
Detroit, MI, USA
dzhu@wayne.edu

***Abstract:*** Urban flooding poses an escalating threat to transportation infrastructure, yet no operational system provides real-time street-level flood depth at centimeter resolution. This paper presents three vision–language models fine-tuned for continuous flood depth estimation from street-level imagery: FloodLlama-Dense, a fully fine-tuned QLoRA baseline; and FloodLlama-MI5 and FloodLlama-MI6, interpretability-guided sparse variants that fine-tune only the top 5 and top 6 causally relevant cross-attention layers identified through mechanistic interpretability analysis, respectively. Training uses a ~610K-image subset of a 2.81M-image synthetic corpus generated in Unreal Engine 5, combining single-vehicle (5 cm steps) and mixed-vehicle (1 cm steps) subsets, spanning 7 vehicle types, 4 weather conditions, and 0–40 cm depths. FloodLlama-Dense achieves MAE = 0.40 cm, RMSE = 1.97 cm, and Acc@5cm = 97.59%. Mechanistic interpretability: combining linear probing, logit lens, centered kernel alignment (CKA), and cross-attention entropy, reveals a two-stage adaptation pattern: layers L13–L22 restructure visual representations, while layer L23 emerges as the layer at which depth first becomes linearly decodable where depth representations first become linearly decodable. FloodLlama-MI5 and MI6 leverage this insight by fine-tuning only 5–6 of the 8 cross-attention layers, achieving 86–88% parameter reduction (6.55–7.86 M vs. 54.4 M trainable parameters) at minimal accuracy cost. On a real-world benchmark, MI6 achieves 98.62% accuracy vs. 86.61% for the published STURM-FloodDepth baseline.

## I. INTRODUCTION

Urban flooding has emerged as one of the most disruptive hazards to modern transportation systems, driven by climate change, aging drainage infrastructure, and rapid urbanization. Even minor inundation can cascade into network-wide failures: percolation analysis of Hurricane Harvey shows that only 2.2% of flood-induced compound failures collapsed 17.7% of a road network's giant component [1], and Wang et al. demonstrate a discontinuous percolation transition where gradual depth increases trigger abrupt network fragmentation [2]. The severity of these disruptions depends sharply on water depth, small differences determine whether a road is passable, restricted, or impassable [3], [4] — yet operational platforms such as Google Maps and Waze still rely on binary closure reports.

The demand for accurate street-level depth is further amplified by emerging vehicle technologies. Electric vehicles (EVs) carry high-voltage battery packs near the underbody and risk thermal runaway at depths above approximately 15–25 cm [5], while autonomous vehicles (AVs) at SAE Level 3–5 routinely exclude flooded roads from their operational design domains because current perception stacks cannot estimate water depth [6]. Critically, even a 1 mm water film can reduce tire–pavement friction by up to 61% at highway speeds [7], and dynamic hydroplaning onsets at water-film thicknesses of just 0.56–1.5 mm [8] — depths that remain undetected by current systems. Both applications require sub-decimeter precision that no infrastructure-free system currently provides.

Vision-based flood depth estimation has progressed through three generations, each with persistent limitations. Reference-object methods [9]-[10] achieve only ∼12 cm MAE and depend on objects not consistently visible. Vehicle-centric detectors [11]-[12] treat depth as coarse discrete categories rather than a continuous variable. A third generation has emerged with large vision–language models (VLMs), which offer a fundamentally different advantage: pretrained on web-scale image–text data, they generalize zero-shot or few-shot to new domains without retraining a task-specific detector or curating reference-object catalogs. Most recently, large multimodal models such as GPT-4V [13] and FloodVision [14] bring zero-shot reasoning but still produce MAE above 8 cm [15] and, as proprietary black-box systems, preclude the interpretability required for safety-critical deployment. To our knowledge, no prior work has fine-tuned an open-source VLM for continuous centimeter-resolution depth estimation, nor dissected the mechanism by which such a VLM acquires this task.

This paper addresses both gaps. We construct a 2.81M-image synthetic flood corpus using Unreal Engine 5 combining single-vehicle (5 cm steps) and mixed-vehicle (1 cm steps) subsets, of which ~610K images are used for fine-tuning the LLaMA 3.2-11B Vision with QLoRA to produce FloodLlama-Dense. We then apply mechanistic interpretability analysis — linear probing [16], login lens [17], LoRA Adapter Weight analysis, centered kernel alignment (CKA) [18], and cross-attention entropy [19] — building on the broader mechanistic interpretability program for deep models [20]. The analysis reveals a two-stage adaptation process: early representational restructuring at layers L13–L22 followed by a sharp phase transition at L23 where depth information first becomes linearly

Corresponding author: Xiaodong Qian (xdqian@wayne.edu)

decodable. Building on these findings, we develop FloodLlama-MI5 and FloodLlama-MI6, which fine-tune layer subsets selected via empirical search informed by the interpretability analysis. Both variants are validated on a 300-image real-world flood test set, outperforming the STURM-FloodDepth baseline by ~10 percentage points.

The contributions of this work are: (1) A 2.81M-image synthetic flood corpus spanning 7 vehicle types, 4 weather conditions, and 0–40 cm depths, combining a 1.73M single-vehicle subset (5 cm steps) and a 1.08M mixed-vehicle subset (1 cm steps); ~610K images are drawn for fine-tuning, with quantitative sim-to-real fidelity assessment.; (2) FloodLlama-Dense, the first open-source VLM fine-tuned for continuous flood depth regression (MAE = 0.40 cm, Acc@5cm = 97.59%); (3) the first mechanistic interpretability analysis of cross attention layers for a flood depth model, identifying layer L23 as the depth-encoding leverage; and (4) FloodLlama-MI5 and MI6, achieving 86–88% parameter reduction while outperforming the published STURM-FloodDepth baseline on real-world data.

## II. RELATED WORK

### A. Flood Depth Estimation from Imagery

Vision-based flood depth estimation has evolved through three paradigms. Reference-object geometric methods estimate depth by comparing the visible portion of a partially submerged object against its known dimensions [9]-[10], but fail when the reference object is absent or occluded. Object-detection methods using Mask R-CNN or the YOLO family [11]-[12] detect vehicles and classify submersion levels (tire-, bumper-, door-level) but treat depth as a discrete category. STURM-FloodDepth [21] extends this into an end-to-end pipeline but retains discrete categories. Zero-shot large multimodal models such as GPT-4V [13] and FloodVision [14] bring open-ended reasoning but report MAE above 8 cm [15]. To our knowledge, no prior work has fine-tuned an open-source VLM for continuous centimeter-resolution depth regression.

### B. Parameter-Efficient Fine-Tuning of VLMs

VLMs such as CLIP [22] and LLaMA 3.2 Vision [23] couple a pretrained vision encoder with a language decoder via cross-attention. LoRA [24] injects low-rank weight updates into selected linear projections, and QLoRA [25] combines 4-bit NF4 quantization with LoRA for single-GPU adaptation of 11B-class models. Transportation applications have focused on broad task evaluation [26] and isolated problems such as pavement damage detection [27], but the which-layers-to-fine-tune question is typically resolved by heuristic rules. We treat this as a measurement problem and answer it with mechanistic interpretability.

### C. Mechanistic Interpretability of Multimodal Models

Mechanistic interpretability seeks to explain how a model computes its outputs—tracing predictions back to the weights, layers, and activations that produce them—rather than merely characterizing what it predicts [20]. Established techniques include linear probing for layer-wise information decoding [16], [28], the logit lens [17] for projecting hidden states into vocabulary space [29]-[17], CKA for cross-model representation-similarity analysis [18], and attention-flow methods [19], [30]. To date, these tools are rarely applied jointly to vision–language models for transportation tasks, and almost never used as the design driver for downstream parameter-efficient training. We close this loop by composing all four methods into a composite layer-importance ranking that directly determines which cross-attention layers are tuned in MI5 and MI6.

## III. SYNTHETIC FLOOD DATASET GENERATION

Supervised flood depth estimation at centimeter resolution is bottlenecked by data: real flood imagery with verified depth labels is scarce, dangerous to collect, and impossible to control along experimental factors. Prior synthetic flood datasets exist [31] but are limited in scale (typically thousands of images) and depth resolution (≥10 cm steps), and lack the factorial control over vehicle geometry and weather needed for systematic robustness analysis. We address this with the first large-scale, centimeter-resolution synthetic flood corpus designed for VLM fine-tuning — a 2.81 M-image dataset generated in Unreal Engine 5 [32] that systematically varies vehicle type (7), weather (4), and flood depth (0–40 cm, with the mixed-vehicle subset at 1 cm steps). This follows established practice in autonomous-driving research where photorealistic simulation enables sim-to-real transfer for safety-critical perception [33]-[34].

### A. Simulation Pipeline

The simulation environment replicates a typical urban roadway scene with realistic asphalt textures, building facades, lane markings, and physically simulated floodwater. Vehicle assets are placed at controlled positions, and floodwater depth is varied programmatically, at 5 cm increments for the single-vehicle subset and 1 cm increments for the mixed-vehicle subset. Each scene is rendered under four lighting/weather conditions — Sunny, Afternoon, Night, and Rainy — to capture the visual diversity needed for real-world generalization. Two complementary subsets are generated: a single-vehicle corpus isolating individual vehicle geometry versus flood depth, and a mixed-vehicle corpus introducing occlusion and visual complexity reflecting real-world scenes.

### B. Dataset Composition and Sim-to-Real Assessment

The single-vehicle subset spans 7 vehicle types (BUS, Trailer, VAN as Big; SUV, Truck, Mini-van as Medium; Sedan as Small) at 9 discrete depth levels (0, 5, …, 40 cm) crossed with 4 weather conditions, yielding 1,725,712 labeled images. The mixed-vehicle subset varies depth at 1 cm increments over the full 1–40 cm range, producing 1,085,218 images. The combined corpus totals 2,810,930 images (Table 1).

Table 1. The Combined synthetic dataset summary.

| Dataset | Depth Levels | Vehicle Types | Weather | Total Images |
|---|---|---|---|---|
| **Single vehicle** | 9 discrete (0,5,10,,40 cm) | 7 types | 4 | 1,725,712 |
| **Mixed vehicle** | 40 continuous (1–40 cm) | Multiple per scene | 4 | 1,085,218 |
| **Combined Total** | 0–40 cm | 7 types + mixed | 4 | 2,810,930 |

Sim-to-real fidelity was assessed via pixel-level KS statistics across 600 random samples per domain, following the two-sample formulation [35]. We adopt the following operational interpretation, consistent with reference benchmarks for KS-derived statistical distances [36]: KS < 0.10 indicates a negligible gap (within sampling noise for the present sample size), 0.10–0.20 a small gap, 0.20–0.30 a moderate gap, and > 0.30 a substantial gap. The mixed-vehicle subset achieves substantially better alignment with real flood imagery than the single-vehicle subset (KS brightness: 0.308 vs. 0.445; KS saturation: 0.167 — a small gap — vs. 0.443). Contrast distributions are in the moderate range for both subsets (KS ≈ 0.24–0.33). Figure 1 shows the synthetic data used for flood depth estimation by weather category.

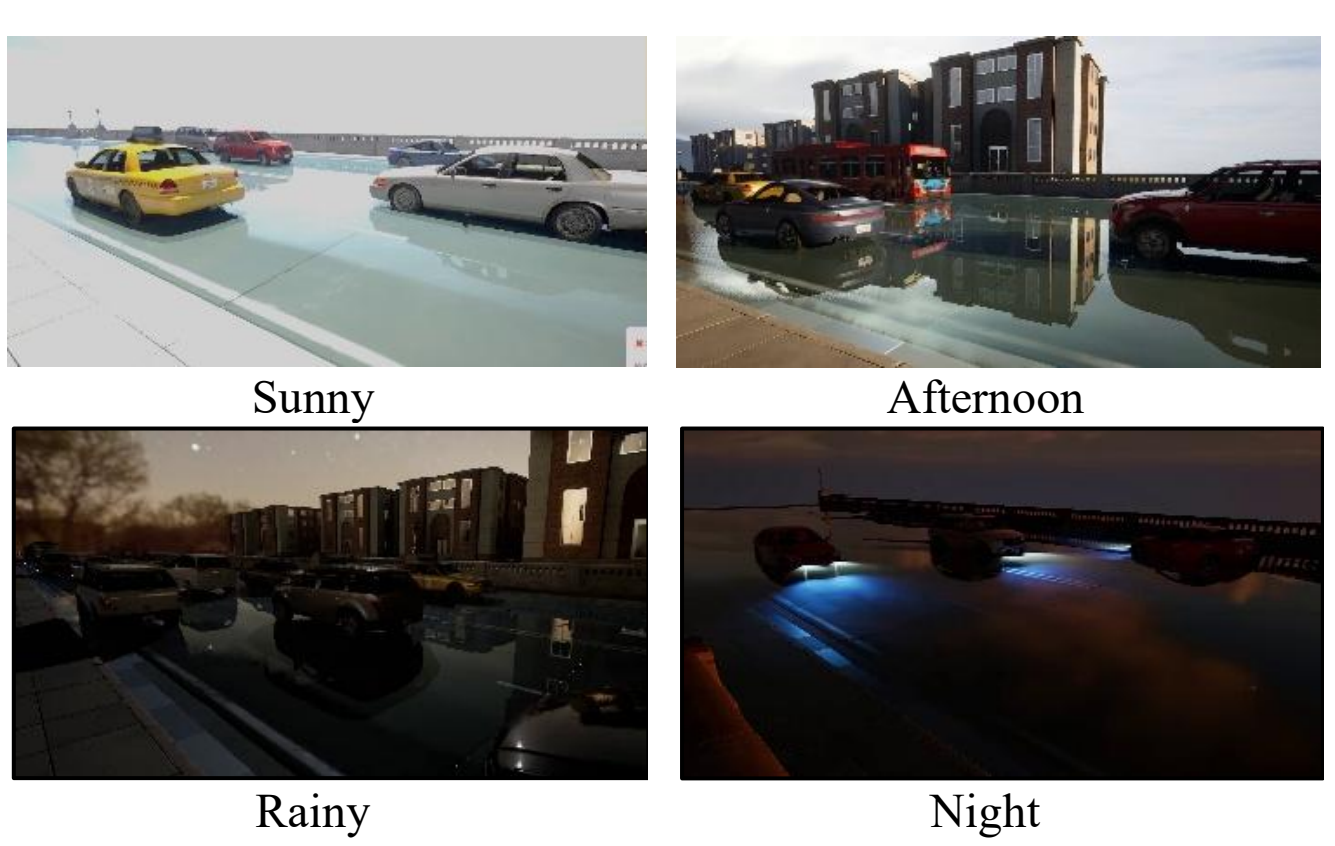

Figure 1. The synthetic images by different weather condition.

## IV. FLOODLLAMA-DENSE: BASELINE FINE-TUNING

We establish a strong baseline by fine-tuning LLaMA 3.2-11B Vision Instruct for continuous flood depth regression. Prior to selection, we benchmarked DeepSeek-VL 3B, Qwen-VL 7B, and LLaVA 13B; LLaMA 3.2-11B achieved the best zero-shot $R^2$ = 0.612. We refer to this fully fine-tuned baseline as FloodLlama-Dense because all eight attention layers receive LoRA adapters.

### A. Model Architecture and QLoRA Configuration

LLaMA 3.2-11B Vision Instruct is a decoder-only multimodal model that couples a frozen ViT-H/14 vision encoder to a 40-layer decoder via a cross-modal adapter. Among those, 32 layers carry self-attention and eight of them (L3, 8, 13, 18, 23, 28, 33, 38) carry cross-attention that fuses the two streams. We adapt the decoder via QLoRA [25] — 4-bit NF4 quantization with LoRA adapters [24] ($r = 16$, $\alpha = 16$) — inserted into the Q/K/V/O projections of all 40 attention modules (32 self-attention + 8 cross-attention). The vision encoder and cross-modal projector remain frozen. This yields 54.4 M trainable parameters (0.49% of the 10.72 B base), enabling adaptation on a single NVIDIA RTX 4090 (24 GB).

### B. Two-Phase Progressive Training

Training proceeds in two phases. Phase 1 fine-tunes on a balanced 340,416-image subsample of the 1,725,712-image single-vehicle subset, drawn via stratified random sampling that retains approximately 1,350 images per (vehicle, depth, weather) cell across the 7 × 9 × 4 factorial design. This subsample size was chosen to balance two constraints: (i) the per-cell count is large enough to support stable gradient estimates for every factor combination, and (ii) the total volume keeps Phase 1 within roughly 12 GPU-hours on a single RTX 4090, consistent with our resource budget. Stratification ensures that every vehicle type, depth level, and weather condition is exposed equally during training, so that the basic vision–language mapping for vehicle geometry and waterline reasoning is not biased toward over-represented cells in the raw corpus. The remainder of the single-vehicle subset is held out for evaluation. Phase 1 fine-tunes for 3 epochs (cosine schedule, peak learning rate $2\times10^{-4}$, AdamW-Paged optimizer, effective batch size 8), establishing the basic vision–language mapping for vehicle geometry and waterline reasoning. Phase 2 adapts the model to mixed-vehicle scenes via cumulative fine-tuning: the mixed-vehicle data are partitioned into nine sequential 30,000-image chunks, and the Phase-1 checkpoint is iteratively updated on each chunk for one epoch at reduced LR of $5\times10^{-5}$. Training loss decreases monotonically from 0.0767 in round 1 to $9.71\times10^{-4}$ in round 9, indicating stable convergence without catastrophic forgetting. The cumulative training strategy is adopted due to the GPU resource limitations.

### C. Performance

Evaluation employs two synthetic test sets drawn exclusively from the held-out partition of the 2.81 M-image corpus, comprising the residual images after exclusion of the approximately 610 K samples allocated to training. Strict separation between training and test partitions is enforced at the scene level to preclude multi-frame leakage. The single-vehicle test set comprises 25,200 images drawn via stratified random sampling at a density of 100 images per cell over the 7 × 9 × 4 (vehicle × depth × weather) factorial design, affording 95% confidence intervals of approximately ± 5 percentage points for per-cell accuracy estimates. The mixed-vehicle test set comprises 16,000 images sampled at 100 per cell over the 40 × 4 (depth × weather) design, yielding approximately 4,000 images per depth category (Low, Medium, High, Very High; Table 3) and supporting reliable depth-stratified inference. After Phase 2 training, FloodLlama-Dense reaches MAE = 0.40 cm, RMSE = 1.97 cm, and $R^2$ = 0.9512 on a 16,000-image mixed-vehicle test set, with exact-match accuracy of 94.92% and Acc@5cm = 97.59% (fewer than 0.35% of predictions deviate by more than 10 cm). The Phase-1 to Phase-2 improvement is dramatic, MAE drops from 5.65 cm to 0.40 cm, a 92.9% reduction, confirming that mixed-vehicle exposure is essential for closing the geometry-occlusion gap (Table 2) and better reasoning. We attribute this primarily to the richer visual signal of mixed-vehicle scenes, where multiple vehicles per scene provide cross-referenceable geometric cues — multiple tire heights, and waterline contacts captured under the same flood depth — allowing the model to triangulate water level rather than depend on a single reference frame. A secondary contribution likely comes from the finer Phase-2 depth resolution (1 cm vs. 5 cm steps), which provides denser supervision in the continuous regression regime. We do not isolate the two effects via a controlled ablation but treat them jointly as the intended Phase 1 to Phase 2 curriculum design. Depth-stratified analysis (Table 3) shows accuracy exceeding 96.8% in the Low, Medium, and High categories, with a modest decline to 93.7% in the Very High (31–40 cm) regime where wheel-arch cues become partially submerged.

Table 2. The Training configuration and overall performance of FloodLlama-Dense.

| Parameter / Metric | Phase 1 (Single-Vehicle) | Phase 2 (Mixed-Vehicle, Final) |
|---|---|---|
| **Training samples** | 340,416 | 270,000 (9 × 30K) |
| **Epochs** | 3 | 9 rounds × 1 epoch |
| **Learning rate** | $2\times10^{-4}$ (cosine) | $5\times10^{-5}$ |
| **Effective batch size** | 8 | 8 |
| **Hardware** | RTX 4090 (24 GB) | RTX 4090 (24 GB) |
| **Testing Samples** | 25,200 | 16,000 |
| **MAE (cm)** | 5.65 | 0.40 |
| **RMSE (cm)** | 7.38 | 1.97 |
| **Acc ±0 cm** | 23.23% | 94.92% |
| **Acc ±5 cm** | 75.42% | 97.59% |
| **$R^2$** | 0.8722 | 0.9512 |

Table 3. The Depth-stratified performance of FloodLlama-Dense.

| Depth Category | Range (cm) | MAE (cm) | Acc ±5 cm |
|---|---|---|---|
| **Low** | 1–10 | 0.18 | 99.1% |
| **Medium** | 11–20 | 0.21 | 98.2% |
| **High** | 21–30 | 0.34 | 96.8% |
| **Very High** | 31–40 | 0.87 | 93.7% |
| **Overall** | 1–40 | 0.40 | 97.59% |

## V. MECHANISTIC INTERPRETABILITY OF FLOODLLAMA-DENSE

FloodLlama-Dense achieves strong accuracy but offers no insight into which of its eight cross-attention layers encode flood depth. This section answers that question through four mechanistic methods — linear probing, logit lens, CKA, and cross-attention entropy — anchored by a preliminary LoRA-weight inspection. All analyses use a held-out subset of 480 synthetic images sampled at 5 per cell from a {3 vehicle size classes (Big / Medium / Small) × 4 weather conditions × 8 depth levels (5, 10, 15, 20, 25, 30, 35, 40 cm, excluding the dry 0 cm condition)} factorial design; activations are collected in fp16 and compared layer-by-layer against the un-tuned LLaMA 3.2-11B Vision-Instruct base model.

### A. LoRA Adapter Weight Analysis

For each linear projection that received a LoRA adapter, we compute the Frobenius norm of the rank-r update, ΔW‖_F = ‖BA‖_F, and aggregate by decoder layer and module type. Two findings dominate. First, five of the eight cross-attention layers — L18, 23, 28, 33, 38 — absorb 2.5–3 times more adaptation than the all-layer average, while the remaining three (L3, L8, L13) absorb at or below the average, localizing depth-related learning to the deeper cross-attention layers within the vision–language fusion pathway. Second, within those high-absorption layers, the largest weight updates concentrate on the query and output projections (q_proj: 513.69, o_proj: 464.98), while the value projection (v_proj: 201.50) changes least; adaptation reshapes how visual features are queried far more than the features themselves. Singular-value decomposition reveals near-rank-1 dominance, consistent with the model learning a small number of dominant weight directions rather than diffusing high-dimensional updates.

### B. Linear Probing: Where Depth is Encoded

We train a ridge regressor on the frozen hidden state of each cross-attention layer to predict ground-truth depth (Table 4). The base model exhibits $R^2 < 0$ at every layer — depth is not linearly decodable from any of its representations. After QLoRA fine-tuning, FloodLlama-Dense (FL) probes remain negative at L3–L18, but the $R^2$ jumps from −0.355 L18 (FL) to +0.513 L23 (FL), a 0.772 absolute increase concentrated in a single layer; probe MAE drops from ∼9 cm to ∼5 cm at the same transition. Downstream layers (L28, L33, L38) maintain comparable probe performance ($R^2 \approx 0.55$), indicating that L23 is the encoding leverage where depth information first becomes available, and that downstream layers preserve and integrate it.

Table 4. The Linear probe results at each cross-attention layer (base vs. FloodLlama-dense).

| Layer | Base $R^2$ | FL $R^2$ | $\Delta R^2$ | Base MAE (cm) | FL MAE (cm) | ΔMAE (cm) |
|---|---|---|---|---|---|---|
| **L3** | −0.301 | −0.317 | −0.016 | 9.81 | 9.86 | +0.05 |
| **L8** | −0.322 | −0.270 | +0.052 | 9.40 | 9.41 | +0.01 |
| **L13** | −0.333 | −0.324 | +0.009 | 9.22 | 9.54 | +0.32 |
| **L18** | −0.305 | −0.355 | −0.050 | 8.95 | 9.68 | +0.73 |
| **L23** | −0.259 | +0.513 | +0.772* | 8.77 | 5.02 | −3.75 |
| **L28** | −0.200 | +0.559 | +0.759 | 8.37 | 4.69 | −3.68 |
| **L33** | −0.232 | +0.553 | +0.785 | 8.34 | 4.58 | −3.76 |
| **L38** | −0.305 | +0.551 | +0.855 | 8.51 | 4.52 | −3.99 |

*Note: FL means FloodLlama-dense. * Phase transition at L23: $\Delta R^2$ = +0.772, probe MAE drops 3.75 cm in one layer.*

### C. Logit Lens: When Language Emerges

The logit lens projects each layer's hidden state through the model's un-embedding matrix to inspect what tokens a layer is "saying." Layers L3–L23 produce incoherent subwords fragments in both base and FloodLlama-Dense. From L28 onward, the base model converges on scene-description vocabulary ("vehicles," "flooded," "image"), while the fine-tuned model produces numeric tokens at L33 ("59," "26," "23") and output-schema tokens at L38 ("depth," "assistant," "flood depth"). The qualitative shift mirrors the probing result: depth-relevant language emerges only after the layer at which depth becomes decodable.

### D. CKA: Representational Drift vs. Task Encoding

CKA measures the similarity of base and FloodLlama-Dense representations on identical inputs, with values near 1

indicating minimal change. Strikingly, the largest representational drift occurs at decoder layers L13–L22 (minimum CKA = 0.395 at L17), not at the depth-encoding layers L23–L38, where CKA remains in the 0.68–0.78 range. Among cross-attention layers specifically, L13 diverges most (CKA = 0.457) while L8 barely changes (CKA = 0.852). This mismatch between representational change and task encoding is the central mechanistic finding: the layers that change the most are not the layers that learn the task. Combined with probing, it points to a two-stage adaptation, early-to-mid layers reorganize visual features, late layers convert the reorganized representation into a numerical depth estimate.

### E. Cross-Attention Entropy and Composite Ranking

Shannon entropy of cross-attention weights measures how diffuse versus focused attention is across image patches. The FloodLlama-Dense shows sharper attention overall (mean entropy 7.15 vs. 7.26 for the base model), with the largest reductions at L13 ($\Delta H = -0.30$), L18 (−0.25), and L28 (−0.20). The layer with the sharpest entropy reduction (L13) is also the layer with the most representational drift but the lowest probe $R^2$ — spatial selectivity is necessary but not sufficient for task encoding.

Table *5* synthesizes four methods into a composite score. L23 ranks first (0.768), confirming its dual role as the largest weight-update layer and the depth-encoding transition layer. L18 ranks second (0.679). Critically, L13 places sixth — its score is driven entirely by CKA drift and entropy focus, with zero probe contribution, the canonical case of restructuring without encoding. This ranking is specific to FloodLlama-Dense's cross-attention layout in LLaMA 3.2-11B Vision Instruct; absolute layer indices (e.g., L23 as the depth-encoding transition layer) will differ for other LVLMs. The composite-ranking procedure itself applies in principle to any VLM exposing identifiable fusion layers, such as Qwen-VL, LLaVA-NeXT, or InternVL.

Table 5. The Composite layer ranking for the eight cross-attention layers of FloodLlama-Dense (built on LLaMA 3.2-11B Vision Instruct), aggregated from four mechanistic interpretability methods.

| Layer | Logit Lens | ‖ΔW‖ Norm | Probe $\Delta R^2$ | 1−CKA | Entropy Focus | Composite | Rank |
|---|---|---|---|---|---|---|---|
| L3 | 0.316 | 0.000 | 0.037 | 0.422 | 0.125 | 0.180 | 8th |
| L8 | 0.970 | 0.054 | 0.112 | 0.000 | 0.520 | 0.331 | 7th |
| L13 | 0.000 | 0.220 | 0.066 | 1.000 | 1.000 | 0.457 | 6th |
| L18 | 0.786 | 0.877 | 0.000 | 0.877 | 0.857 | **0.679** | **2nd** |
| L23 | 0.939 | 1.000 | 0.908 | 0.438 | 0.555 | **0.768** | **1st *** |
| L28 | 0.490 | 0.692 | 0.893 | 0.261 | 0.738 | **0.615** | **3rd** |
| L33 | 0.500 | 0.454 | 0.922 | 0.390 | 0.293 | **0.512** | **4th** |
| L38 | 1.000 | 0.329 | 1.000 | 0.195 | 0.000 | **0.505** | **5th** |

*Note: All metrics min-max normalized to [0, 1] across the eight cross-attention layers; composite is the arithmetic mean. Bold rows indicate the top five layers by composite score.*

## VI. FLOODLLAMA-MI5 AND MI6: INTERPRETABILITY-GUIDED EFFICIENT TRAINING

### A. Layer Selection Rationale

The composite ranking of Table 5 indicates the concentration of depth-relevant adaptation in a subset of the eight cross-attention layers, but not in a way reducible to a single threshold rule. We therefore conducted an empirical search over combinations of the eight cross-attention layers, evaluating each candidate configuration on the validation set. Two configurations consistently outperformed the alternatives and are reported here:

**FloodLlama-MI5** tunes L8, 18, 23, 33, 38 — the minimal selection retaining the encoding leverage (L23), the dominant weight-adaptation layer (L18), and the language-integration layers (L33, L38).

**FloodLlama-MI6** adds L28 to the MI5 set, providing a redundant encoding pathway around L23 to improve robustness when L23's representation is degraded by visual occlusion.

### B. Training Protocol and Parameter Summary

Both variants are trained with the same QLoRA configuration as FloodLlama-Dense (4-bit NF4 base, LoRA r = 16, α = 16, applied to q/k/v/o projections), but with adapters inserted only into the selected cross-attention layers; all other layers remain frozen. The two-phase progressive curriculum is identical to Dense (Phase 1: single-vehicle, Phase 2: nine cumulative mixed-vehicle chunks) with the same learning rates, optimizer, batch size, and hardware. This preserves experimental control, any performance difference is attributable to layer selection alone. Model suffixes (MI5, MI6) denote the number of cross-attention layers selected for fine-tuning (Table *6*).

Table 6. The three-model configuration summary.

| Model | Cross-Attn Layers w/ LoRA | Trainable Params | % of 10.7B | Reduction vs. Dense |
|---|---|---|---|---|
| **FloodLlama-Dense** | All 8 (L3,8,13,18,23,28,33,38) + self-attn in 32 other layers | 54.4 M | 0.49% | — (baseline) |
| **FloodLlama-MI5** | 5 of 8 (L8,18,23,33,38) | 6.55 M | 0.06% | 88% reduction |
| **FloodLlama-MI6** | 6 of 8 (L8,18,23,28,33,38) | 7.86 M | 0.07% | 86% reduction |

***Note:*** *"% of 10.72B" compares against the Llama 3.2 Vision Instruct model*.

## VII. EXPERIMENTAL EVALUATION

We evaluate all three FloodLlama variants across three conditions: (i) synthetic holdout test set, (ii) real-world flood photographs benchmarked against STURM-FloodDepth [21], and (iii) systematically occluded inputs that stress depth-cue redundancy.

### A. Experimental Setup

Synthetic test set: 5,000-image holdout from the mixed-vehicle subset, stratified by depth (1–40 cm at 1 cm intervals), balanced across four weather conditions. Real-world test set: 300 real flood photographs from the STURM-FloodDepth benchmark [21], manually annotated with reference-object-derived ground-truth depth (±2 cm typical uncertainty), covering a range of vehicle types and flood conditions. Metrics: MAE (cm), RMSE (cm), $R^2$, and within-tolerance accuracies at 0, 5, and 10 cm bands. For the STURM comparison we additionally report per-level accuracy (Level 0: no flood; Level 1: depth ≤30 cm; Level 2: depth >30 cm) and macro-F1, since STURM produces discrete categorical outputs. Because the STURM-FloodDepth benchmark provides only discrete categorical labels rather than continuous depth measurements, continuous regression metrics (MAE, RMSE) cannot be computed on this test set. Comparison against STURM is therefore restricted to the categorical metrics it natively supports.

### B. Three-Way Model Comparison (Synthetic)

Table 7 reports headline metrics on the synthetic holdout. FloodLlama-Dense achieves the lowest absolute error (MAE = 0.40 cm, Acc@5cm = 97.59%), as expected given its 8.3× larger trainable footprint. FloodLlama-MI5 doubles the MAE in absolute terms (0.40 cm to 0.80 cm) but remains well within the 5 cm operational tolerance relevant for vehicle traversability decisions (Acc@5cm = 95.1%), while training 88% fewer parameters. FloodLlama-MI6 narrows the gap slightly with its additional L28 adapter (MAE = 0.78 cm, Acc@5cm = 95.8%), though MAE remains close to double that of Dense. Depth-stratified performance is consistent across all three models: lowest error in the 1–20 cm range, modest degradation above 30 cm. Interpretability-guided layer selection therefore preserves operational accuracy within the ±5 cm tolerance band at a fraction of the training cost, accepting a roughly 2× absolute MAE penalty relative to the dense baseline.

Table 7. The three-way model comparison on the synthetic holdout test set.

| Model | MAE (cm) | RMSE (cm) | $R^2$ | Acc ±0cm | Acc ±5cm | Trainable Params |
|---|---|---|---|---|---|---|
| **FloodLlama -Dense** | 0.40 | 1.97 | 0.951 | 94.9% | 97.6% | 54.4 M |
| **FloodLlama -MI5** | 0.80 | 2.64 | 0.931 | 88.1% | 95.1% | 6.55 M |
| **FloodLlama -MI6** | 0.78 | 2.51 | 0.937 | 89.4% | 95.8% | 7.86 M |

### C. Benchmark Against STURM-FloodDepth

We compare the three efficient variants — FloodLlama-MI5, FloodLlama-MI6 and FloodLlama-Dense — against STURM-FloodDepth [21] on the real-world test set, using three prompt styles: Simple, Detailed, and Chain-of-Thought (CoT).. FloodLlama continuous predictions are discretized into STURM's categorical levels for direct comparison. Table 8 and Table 9 report per-level accuracy on real-world and synthetic data respectively. FloodLlama-Dense, MI5 and MI6 substantially outperform STURM (except MI5-Simple for Level 2) on overall accuracy: FloodLlama-MI6 with CoT achieves 98.62% vs. 86.61% for STURM — a 12.01 pp improvement — with 100.00% Level 2 accuracy (vs. 88.27% for STURM). All variants outperform STURM on overall accuracy and on nearly every category, with the single exception of MI5-Simple on Level 2, demonstrating that interpretability-guided sparsity does not sacrifice real-world generalization. Across all real-world comparisons, 95% confidence intervals span ±1.32% – ±3.85% (overall accuracy, n=300), with non-overlapping intervals between all FloodLlama variants and STURM confirming statistical reliability of the reported gains. On the synthetic benchmark, confidence intervals are substantially tighter (±0.37% – ±1.38%, n=5,000), reflecting the larger test set size.

Table 8. The Level-wise accuracy on real-world data.

| Model / Prompt | Overall, Acc. | Level0 Acc. | Level1 Acc. | Level2 Acc. |
|---|---|---|---|---|
| **STURM (Baseline)** | 86.61% | 95.64% | 77.64% | 88.27% |
| **FloodLlama-MI5 — CoT** | 97.23% | 98.84% | 94.76% | 97.85% |
| **FloodLlama-MI5 — Detailed** | 96.28% | 98.02% | 96.74% | 97.58% |
| **FloodLlama-MI5 — Simple** | 93.41% | 98.67% | 98.58% | 82.97% |
| **FloodLlama-MI6 — CoT** | 98.62% | 98.18% | 97.17% | 100.00% |
| **FloodLlama-MI6 — Detailed** | 96.44% | 99.59% | 98.02% | 92.03% |
| **FloodLlama-MI6 — Simple** | 95.19% | 99.17% | 97.03% | 89.70% |
| **FloodLlama-Dense — CoT** | 98.13% | 98.93% | 95.46% | 100.00% |
| **FloodLlama-Dense — Detailed** | 97.54% | 99.17% | 97.74% | 95.70% |
| **FloodLlama-Dense — Simple** | 97.20 % | 97.62% | 96.11% | 97.87% |

Table 9. The level-wise accuracy on synthetic data.

| Model / Prompt | Overall, Acc. | Level0 Acc. | Level1 Acc. | Level2 Acc. |
|---|---|---|---|---|
| **STURM (Baseline)** | 51.89% | 84.47% | 5.79% | 64.87% |
| **FloodLlama-MI5 — CoT** | 90.97% | 89.50% | 93.70% | 89.70% |
| **FloodLlama-MI5 — Detailed** | 92.17% | 85.00% | 98.00% | 93.49% |
| **FloodLlama-MI5 — Simple** | 95.73% | 96.50% | 97.20% | 93.52% |
| **FloodLlama-MI6 — CoT** | 98.17% | 99.10% | 96.60% | 98.80% |
| **FloodLlama-MI6 — Detailed** | 94.83% | 94.40% | 92.00% | 98.10% |
| **FloodLlama-MI6 — Simple** | 90.97% | 98.40% | 82.70% | 91.80% |
| **FloodLlama-Dense — CoT** | 97.48% | 96.42% | 97.20% | 98.82% |
| **FloodLlama-Dense — Detailed** | 96.86% | 97.21% | 95.32% | 98.06% |
| **FloodLlama-Dense — Simple** | 95.88% | 96.90% | 95.90% | 94.85% |

Per-level results reveal a prompt-induced calibration effect: the Simple prompt yields the highest Level 1 accuracy (MI5: 98.58%; MI6: 97.03%) but the lowest Level 2 accuracy (MI5:

82.97%; MI6: 89.70%), biasing the model toward moderate-depth predictions and misclassifying deeper cases downward into Level 1. Detailed and CoT prompts shift the decision boundary closer to the 30 cm threshold. We recommend CoT for safety-critical deployment, since it combines the highest overall accuracy with the best Level 2 performance, and under-predicting deep flood is operationally more dangerous than under-predicting moderate flood. This suggests that explicit reasoning prompts act as a calibration mechanism for VLM-based depth classifiers.

## VIII. DISCUSSION AND CONCLUSION

### A. Discussion

A notable finding is that fine-tuning a VLM for a regression task does not distribute change uniformly across layers. Layers L13–L22 undergo substantial representational restructuring (low CKA) without contributing to depth encoding (zero probe $\Delta R^2$), while layers L23–L38 retain representations comparable to the base model (CKA > 0.68) but newly support linear decoding ($R^2 > 0.5$). The transition is sharp, concentrated almost entirely at L23, rather than gradual. This restructure-then-encode pattern is consistent with probing studies in language-only models showing that task-relevant features emerge progressively across layers [37], with task-specific knowledge concentrating in middle-to-late feed-forward layers [38].

This has practical implications for Parameter Efficient Fine-tuning (PEFT) methodology. Standard fine-tuning treats layer selection as a hyperparameter, typically tuning all attention layers or the last k layers. Our results suggest that measurement — probing, CKA, and logit lens — offers a complementary approach with a fixed cost (~40 forward passes plus 40 closed-form ridge probes), independent of the number of candidate subsets and considerably smaller than the combinatorial search space subset methods must navigate. The 86–88% parameter reduction by MI5 and MI6 at minimal accuracy cost indicates that interpretability-guided PEFT may generalize to other VLMs adapted for specialized regression tasks beyond the flood-depth setting.

### B. Limitations and Future Work

Three limitations bound these conclusions. First, the analysis is conducted on a single VLM architecture (LLaMA 3.2-11B Vision); the L23 leverage and restructure-then-encode pattern may shift with different vision encoders, cross-attention layouts, or model scales. Second, the 300-image real-world test set under-represents three long-tail conditions: night-time and low-light scenes, extreme depths above 30 cm (Level 2 cases), and flood scenes lacking vehicle reference objects. Third, depth-stratified errors above 30 cm confirm that reliance on wheel-arch cues becomes brittle when those features submerge, reinforcing the need for reference objects beyond vehicles. Future work will extend the interpretability protocol to other VLM architectures (Qwen-VL, LLaVA-NeXT), expand the real-world benchmark along these three under-represented dimensions, apply the method to other transportation regression tasks (vehicle speed estimation, road-surface friction), and explore whether composite layer rankings can be optimized end-to-end.

### C. Conclusion

This paper presented FloodLlama-Dense, FloodLlama-MI5, and FloodLlama-MI6 — three VLMs fine-tuned for centimeter-resolution flood depth on a ~610K-image training subset drawn from a 2.81M-image synthetic corpus (single-vehicle at 5 cm steps, mixed-vehicle at 1 cm steps) generated in Unreal Engine 5. Mechanistic interpretability of the fully fine-tuned baseline revealed a two-stage adaptation pattern in which layer L23 acts as the depth-encoding transition layer and earlier layers (L13–L22) restructure visual representations without contributing to encoding. FloodLlama-MI5 and MI6 tune only the interpretability-identified layers, reducing trainable parameters by 86–88% while outperforming the published STURM-FloodDepth baseline (98.62% vs. 86.61% on real-world data). The results establish mechanistic interpretability as a practical tool for guiding PEFT of VLMs in transportation perception tasks.